\documentclass{IOS-Book-Article}

\usepackage{mathptmx}
\usepackage{soul}\setuldepth{article}
\usepackage{ucs}
\usepackage[utf8x]{inputenc}
\usepackage[T1]{fontenc}
\usepackage[english,greek]{babel}
\usepackage{graphicx}
\usepackage{amssymb}
\usepackage{amsmath}
\usepackage{amsthm}
\usepackage{bbold}  
\usepackage{stackengine} 

\theoremstyle{thmstyleone}%
\theoremstyle{thmstylethree}%
\newtheorem{definition}{Definition}%

\def\hb{\hbox to 11.5 cm{}}

\begin{document}
\selectlanguage{english}

\pagestyle{headings}
\def\thepage{}
\begin{frontmatter}              

\title{A model of interaction semantics}
\author[A]{\fnms{Johannes} \snm{Reich}\orcid{0000-0002-0378-853X}}
\address[A]{SAP SE, Dietmar-Hopp-Allee 16, 69190 Walldorf, Germany}

\markboth{}{June 2023\hb}

\begin{abstract}
Purpose: The purpose of this article is to propose, based on a model of an interaction semantics, a certain understanding of the ''meaning'' of the exchanged characters within an interaction. 

Methodology: Based on a model of system interaction, I structure the model of interaction semantics similar to the semantics of a formal language: first, I identify adequate variables in my interaction model to assign values to, and second, I identify the interpretation function to provide meaning. Thereby I arrive at a model of interaction semantics which, in the sense of the late Ludwig Wittgenstein, can do without a 'mental' mapping from characters to concepts. 

Findings: The key findings are a better understanding of the tight relation between the informatical approach to model interactions and game theory; of the central 'chicken and egg' problem, any natural language has to solve, namely that to interact sensibly, we have to understand each other and to acquire a common understanding, we have to interact with each other, which I call the 'simultaneous interaction and understanding (SIAU)' problem; why ontologies are less 'semantic' then their proponents suggest; and how 'semantic' interoperability is to be achieved.

Value: The main value of the proposed model of interaction semantics is that it could be applied in many different disciplines and therefore could serve as a basis for scientists of natural sciences and humanities as well as engineers to understand each other more easily talking about semantics, especially with the advent of cyber-physical systems.

\end{abstract}

\end{frontmatter}

\begin{center}{\it I dedicate this article to Bernd Finkbeiner}\end{center}
%
\section{Introduction}
%

In our normal live, our intuitive concepts of understanding and meaning prove to be extremely powerful every day. Meaning seems to be {\it the} key concept for our understanding of the function of our natural language. Intuitively, we say that the signs exchanged in a conversation have a meaning that is to be understood by the participants. What do we mean by that? What are the benefits to talk this way? 

Accordingly, the question what comprises this concept of meaning, or semantics, puzzles philosophers since ancient times, accompagnied by linguists, sociologists, computer scientists, etc. starting in the 19th century. 
Actually, the ancient Greek word ''{\selectlanguage{greek}σημαντικός}'' (semantikos) is usually translated as ''significant'' or ''meaningful''. One key issue of natural language semantics seems to be indeed to separate the relevant from the irrelevant (be it appropriate or not) in sharp contrast to formal semantics. How comes that? 

My contribution in this article is a proposition of a certain model of semantics based on a certain model of system interactions. As this is a conceptual article about a model and not a theory, I do introduce formal definitions to make my thoughts as concrete as possible, but I do not derive any major theorem. 

My hope is that the proposed model of interaction semantics could serve scientists of natural sciences and humanities as well as engineers to understand each other more easily talking about semantics, especially with the advent of cyber-physical systems that are just on the doorstep. 

To make this proposition convincing, I have to show that the proposed concept of semantics does indeed convey the essential aspects of what we colloquially denote as semantics. Additionally, it would be desirable from the reader's perspective if I succeeded in explaining to her the advantages of this way of looking at semantics which are to better understand 

\begin{enumerate}
\item the tight relation between the informatical approach to model interactions and game theory;
\item the central 'chicken and egg' problem, any natural language has to solve, namely that to interact sensibly, we have to understand each other and to acquire a common understanding, we have to interact with each other, which I call the 'simultaneous interaction and understanding (SIAU)' problem;
\item why ontologies are less 'semantic' then their proponents suggest; and
\item how 'semantic' interoperability is to be achieved.
\end{enumerate}

The structure of the rest of the article is as following:
In section \ref{s_setting_the_stage}, I sketch the concepts of semantics and the contribution of informatics. In section \ref{s_description_of_systems}, I introduce the protocol concept as a model of interaction that depends on information exchange, that is, on the identical naming of ''exchanged'' characters. Next, in section \ref{s_interaction_semantics}, I first introduce the decision concept to make the protocol executable in a functional sense, despite its nondeterminism. Secondly, I define a fulfillment relation where the assignment of the set of alphabets to a protocol makes it consistent or inconsistent. This approach requires the definition of an interpretation function of the protocol and its constituents, namely its characters. This interpretation is the execution of the protocol and the interpretation of the characters during execution results in their meaning. 
Section \ref{s_other_work} provides a brief overview of the relevant work of others. I conclude this article in the final section \ref{s_discussion} with a summarizing discussion and some speculations.

\section{Setting the stage} \label{s_setting_the_stage}
Unfortunately -- or fortunately --, because of its deep historical roots and its relevance for a plethora of different branches of science, it is not possible to give an even superficial outline of the overall development of the field. I will limit my introduction to just sketch the most relevant developments for the ideas delineated in this article. 

In {\selectlanguage{greek}Περὶ Ἑρμηνείας} (Peri Hermeneias) \cite{Aristoteles_Organon_Bd2_en}, Aristoteles introduced the semantics of a sign as a ternary relation between the sound or word, the thing designated and the mental conception triggered by the sound. In the 19th century, Charles S. Pierce took up this idea again \cite{Peirce1894_What_is_a_Sign} and it became quite influential with the so-called ''semiotic triangle'' \cite{OgdenRichards1923_Meaning}. 

The topicality of these considerations can be seen in the current German DIN/DKE standardisation roadmap Industry 4.0 \cite{DKE2023_NormungsroadmapI40v5}, which refers directly to this ternary relationship, locating things in the ''real world'' and symbols and terms in the ''information world''. Along these ideas, the IEC \cite{IEC2019_WP_SemInterop} then positions informatical ontologies as ''semantic models'' that supposedly endow IT applications with semantic content and simplify semantic interoperability. 

However, this approach is systematically flawed. It tries to explain a complex concept, the one of 'semantics', with several other concepts that are at least as complex by themselves, like 'mental conception', 'real vs. information world', etc. It even somehow requires the entities of the different 'worlds' to map onto each other without providing a model for this mapping. And it does not explain why people with an obvious very different understanding of a concept still can get along together quite well. For example, the 10-year-old son understands his mother, a physics professor with her specialty in quantum many-particle systems, sufficiently well when she asks him, who comes home too early from school on a hot summer day, if he got heat-free because the temperature was too high.   

It was the late Ludwig Wittgenstein, who opposed precisely the notion that every word ''has'' a meaning in the sense that the meaning is the ''object'' for which the word stands. Instead, he proposed to understand the meaning of the words of our language simply as their ''use in language'' \cite{Wittgenstein1953_PI}.   

In this sense language becomes an interactive game in which a competent speaker must know the common rules of word usage. David Lewis \cite{Lewis1969_Convention} successfully took up this line of thought and focused on the role of social conventions for the emergence of meaning in our language expressions. He viewed mutual understanding in an exchange of characters as a coordination problem and introduced signaling games as an analytical instrument. He identified the signal's relation to either the state of the sender ({\it ''signal-that''}) or to the action of the receiver ({\it ''signal-to''}) as its meaning. 

Vincent P. Crawford and Joel Sobel stressed the importance of aligned preferences \cite{CrawfordSobel1982}. Reinterpreting the benefit as an evolutionary selection advantage, Karl Warneryd \cite{Warneryd1993_Cheap_talk} showed that such ''conventions'' arises naturally in evolutionary settings, based on the concept of evolutionary stable strategies as developed by John M. Smith and George R. Price \cite{MaynardSmithPrice1973}. 

The interactive aspect of language was strongly emphasized by Herbert P. Grice \cite{Grice1989_Studies} by noting that to understand an utterance is to understand what the utterer intended to convey - and that what has been traditionally understood as its ''meaning'' is only loosely related to that. Hai Zhuge \cite{Zhuge2010_InteractiveSemantics} also views language as an interaction tool and thus emphasizes the essential role of interaction for semantics. Kasia M. Jaszczolt \cite{Jaszczolt2016_Meaning} proposed that to understand the concept of meaning one has to investigate {\it ''not the language system and  its relation to context but principally the context of interaction itself, with all its means of conveying information''} (pp.12-13).

And now, what is the genuine informatical contribution to our concept of interaction semantics? Is it directed towards the line of thought of Brian Skyrms, who identifies the information content of a signal in how it affects the probabilities of states and actions \cite{Skyrms2010_Signals} (p.34). He relates the quantity of information of a signal to the a priori probabilities of either sender state or receiver action. Thereby he arrives at different information contents of a single message, depending on its assumed relation.

I don't think so. According to my understanding, the information content of a message is uniquely defined. The weakness of the approaches of David Lewis, Brian Skyrms and others is not to distinguish precisely enough between the different languages involved and not to become explicit enough what the inner relation between such entities like signal, state, action and dicisions are. 

From an engineering perspective we use our engineering language to talk about the interaction language. And only the latter is our language of interest. To distinguish it from the others, I name it ''I-language'' as proposed by \cite{BangemannDiedrichReich2016}\footnote{Please note that this term was already used differently by Noam Chomsky to distinguish between the ''I''(intensional)-language, referring to the internal linguistic knowledge, and the ''E''(extensional)-language, referring to the observable language people actually produce \cite{Araki2017_Chomsky}.}. 

In the following, I will dwell on the genuine contribution of informatics to the topic of the semantics of the I-language.

\subsection{Informatics and semantics} \label{ss_informatics_and_semantics}
Ralph V. L. Hartley \cite{Hartley1928}, Claude E. Shannon \cite{Shannon1948, Shannon1949_Noise} and others had the pioneering idea, not to consider the states of the world as ''physical'' locations, velocities, pressures or voltages, but to focus on the distinguishable. To continue to be able to talk about the same states as before, they had to introduce an additional alphabet with a unique character for each distinguishable physical state value into their engineering language: information emerged.  A character in the sense of information theory is a unique name in our engineeering language for a physical state value that can be distinguished from all the other values this state can take.
 
Equipped with this concept of information, we can now describe the essence of what happens between two ''communicating'' systems in the sense that a ''receiver'' system somehow reproduces a distinguishable state value that a ''sender'' system has provided beforehand while both systems remain structurally invariant. If we agreed beforehand on using the same character in our engineering language to denote this reproduced state value in both systems we can speak about ''information transport''.

Remarkably, Claude E. Shannon already related information processing to the field of semantics, and valued the meaning of the transported information as being irrelevant for solving the engineering problem how best to describe its tranport, i.e. communication. He wrote in the introduction to \cite{Shannon1948}: {\it''The fundamental problem of communication is that of reproducing at one point either exactly or approximately a message selected at another point. Frequently the messages have meaning; that is they refer to or are correlated according to some system with certain physical or conceptual entities. These semantic aspects of communication are irrelevant to the engineering problem.''}

Thereby information theory separates information transport from information processing: information transport is about reproducing distinguishable state values {\it between} systems -- enforcing the discussed naming conventions for the necessary characters in our engineering language description of multiple systems. And information processing is about relating arbitrary state values to other arbitrary state values on the base of these naming conventions, happening {\it within} systems.

From an informatics point of view, it is clear, that during interaction, {\it only} information is transported, and {\it not} meaning. So, if we are looking for something like the meaning of the transported information we have to focus on its processing. In this sense, many computer scientists intuitively qualify many concepts related to information processing, such as the layering of an application or the data concept, as being ''semantic'' \cite{Reich2022_VI_1}. In cryptology, one speaks of ''semantically'' secure encryption in precisely this sense if information about the plaintext can only be extracted from the ciphertext with negligible probability \cite{KatzLindell2015}.

Thus, I identify attributing meaning to information with its processing. So, someone with the expectation, that the model of interaction semantics ought to show a way how to implement 'meaning' in computer interactions will be either disappointed or relieved, learning that within this model you cannot process information without attributing meaning and looking at the meaning of any exchanged characters will not add anything special to the interaction within it is exchanged. 

Wouldn't it be convincing to be on the right track, if our main idea to identify the attribution of meaning to information with its processing, would show a tight conceptual connection to the well accepted approach of mathematics, to talk about the semantics of formal languages? This is indeed the case. The definition of the semantics of a formal language is an 'ordinary' mathematical concept that consists of two steps: first to assign values to variables and second to provide meaning by an interpretation function. I delineate it in appendix \ref{ss_concept_of_meaning_in_formal_languages} for the propositional calculus.

The first key idea of this contribution is to apply this approach also to the interaction itself in the line of thought of Gerard Holzmann \cite{Holzmann1991} who already noted the similarities between protocols and natural language. Using the concept of the semantics of formal languages for the semantics of interactions, we get a good justification to also talk about meaning in interactions.

%
\section{The description of systems and their interactions} \label{s_description_of_systems}
%
To describe the interaction of systems, I follow the approach of \cite{Reich2022_VI_1,Reich2023_VI_2}. A system separates an inner state from the state of the rest of the world, the environment. A state in this sense is a time dependent function, taking a single out of a set of possible values, the alphabet $A$, at a given time\footnote{This state concept actually captures the perspective of classical physics. In quantum physics, a state is a more complex concept.}.

How does this separation occur? These time-varying values are not independent, but some of them are somehow related. If this relation is a function, then based on its uniqueness property, this function -- I call it ''system function'' -- allows the identification of a system by identifying the state functions of a system and attributing them their input-, output-, or inner character. Such a functional relation logically implies causality and a time scale.  

Depending on the class of system function or time, different classes of systems can be identified. For the purpose of our investigation, resting on information theory, I will focus on discrete systems. I describe the behavior of a (possible projection of a) discrete system by input/output transition systems (I/O-TSs) of the following form\footnote{Another name in the literature is ''transducer'' \cite{Sakarovitch2010}, because this machine translates a stream of incoming characters into a stream of outgoing characters.}: 

\begin{definition} \label{def_IO-transition_system_mit_epsilon}
An input/output transition system (I/O-TS) ${\mathcal A}$ is given by the tuple ${\mathcal A} = (I, O, Q, (q_0, o_0), \Delta)$ with $I$ and $O$ are the possibly empty input and output alphabets and $Q$ is the non empty set of internal state values, $(q_0, o_0)$ are the initial values of the internal state and output and $\Delta_{\mathcal A} \subseteq I^\epsilon \times O^\epsilon \times Q \times Q$ is the transition relation describing the behavior of a discrete system or its projection, where $\epsilon$ is the empty character and $I^\epsilon = I \cup \{\epsilon\}$ and $O^\epsilon = O \cup \{\epsilon\}$
\end{definition}

The timing behaviour of the system is implicitely contained in the transition relation by attributing the input character and the start state to some time $k$ and the output character and the target state to its successor time $k+1$. Instead of writing $(i_k,o_{k+1},p_k,p_{k+1}) \in \Delta$, I also write $p_k \stackrel{i_k/o_{k+1}}{\rightarrow} p_{k+1}$.

An {\it execution fragment} of an I/O-TS is a sequence of 3-tuples, listing the values that the input, output and inner state functions of the corresponding system have at the considered times: $(i_0, o_0, p_0), (i_1, o_1, p_1), \dots, (i_n, o_n, p_n)$. I name an execution fragment which starts with an initial state a {\it ''run''}.

\subsection{System interactions \label{ss_system_interaction}}
\begin{figure}[htbp]
  \begin{center}
    \includegraphics[width=4cm]{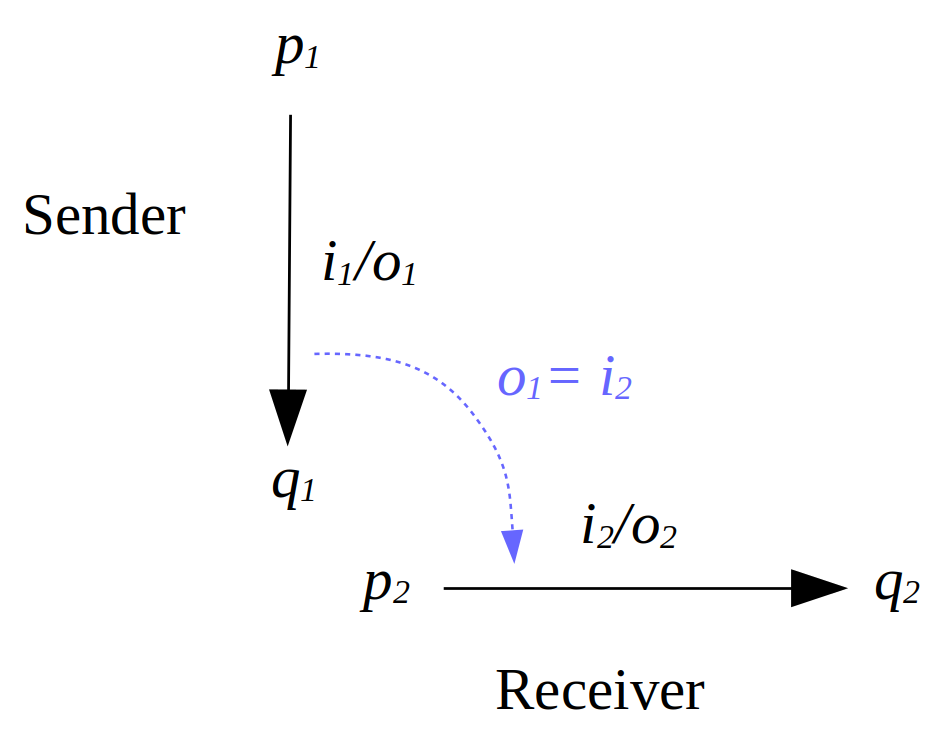}
 \end{center}
\caption{Interaction between two systems in which the output character of a ''sender'' system is used as the input character of a ''receiver'' system.} \label{fig_coupling_by_interaction}
\end{figure}

In our model, interaction simply means that information is transmitted. Accordingly, the description of interaction is based on the use of equally named state values, i.e. 'characters' in the sending and receiving systems such that the state values of an output component of a transition of a 'sender' system are reproduced in the input component of the 'receiver' system and serve there as input of a further transition\footnote{This is quite different to the ''standard'' three-tuple transition systems of ordinary automata theory with a single transition label in the form of $p \stackrel{a}{\rightarrow} q$. Here the label somehow represents the transiion and not any input or output. Accordingly, the coupling of different transitions is usually achieved by synchronizing transitions with identical labels.} (see Fig. \ref{fig_coupling_by_interaction}). 

I call such a state function that serves as output as well as input of two systems a ''Shannon state function'' or ''Shannon state''. 

\subsection{Protocols}\label{ss_protocols}
In the following I focus on systems which interact with multiple other systems in a stateful and nondeterministic way. Their significance is indicated by the many names coined for these kind of systems in the literature like ''processes'' (e.g. \cite{Milner1992}), ''reactive systems'' (e.g. \cite{HarelPnueli1985_Reactive}), ''agents'' (e.g. \cite{Poslad2007_Protocols}) or ''interactive systems'' (e.g. \cite{Broy2010_Logical}). I think the term ''interactive systems'' is most appropriate. Their interactions can be described by protocols \cite{Holzmann1991}.

In deterministic interactions a natural ''purpose'' in the sense of a consequence of composition is simply the creation of the resulting super-system \cite{Reich2010,Reich2021_Komposition_Interoperabilitaet}. However,  in nondeterministic interactions we need an additional criterion for success. By adding a ''success model'', represented by an additional acceptance component $Acc$, we get from I/O-TSs to I/O automata (I/O-Aa). 

For finite calculations with some desired final state value, $Acc$ is the set of all final state values. For infinite calculations of a finite automaton there are differently structured success models. One of them is the so-called Muller acceptance, where the acceptance component is a set of subsets of $Q$, i.e. $Acc_\text{\it Muller} \subseteq \wp(Q)$. An execution (see below) is considered to be successful whose finite set of infinitely often traversed state values is an element of $Acc$  (e.g. \cite{DBLP:conf/dagstuhl/Farwer01}).

To compose different I/O-Aa, we have to add the information which output and input components of our alphabets actually represent the same Shannon state function. Thus we have to complement the simple product automaton of all I/O-Aa with this information and specify the execution rule accordingly. Because of this extra component, the resulting automaton is no longer an I/O-A, although it is quite similar. Its execution rule pays special attention to the character exchange unknown to an I/O-A. This coupling actually restricts the transition relation compared to the uncoupled product I/O-A.

To simplify our further considerations, I assume any I/O-character $c\in I \cup O$ to have at most one component unequal to the empty character $\epsilon$. 

\begin{definition} \label{def_protocol}
A protocol ${\mathcal P} = (A, C)$ is a pair of a set $A=\{{\mathcal A}_1, \dots, {\mathcal A}_n\}$ of I/O-Aa, and a set of index pairs $C$ indicating which output component relates to which input component of $A$ in the sense of a Shannon state. I name the ${\mathcal A}_i$ the ''roles'' of the protocol.

The execution rules are as following with the convention that the current values of $i$, $o$ and $q$ are indicated by a $*$ and the values calculated in the current step by a $+$: 

\begin{enumerate} 
\item {\bf Initialization (time $j=0$):} $(q^*, o^*) = (q_0, o_0)_{\mathcal P}$.

\item {\bf Loop:} \label{calc_I/O-A_loop} Determine the set $T^*$ of all possible transitions for the current state $q^*$. If $T^*$ is empty, end the calculation.

\item {\bf Determine input character $i^*$:} \label{determine_input_marks} Proceed in the following sequence:
\begin{enumerate}
  \item If the current output character $o^*\in O_{\mathcal P}$ has the value $v\neq\epsilon$ in its $k$-th component, and this component is part of a Shannon channel $c=(k,l)$ to the input component $1 \leq l \leq n_I$, then set the $l$-th component of $i^*$ to $v$. 
  \item Otherwise, select one of the possible input characters as $i^*$ for $q^*$ (if some element of $T^*$ is a spontaneous transition, the selection $i^* = \epsilon$ is also allowed).
  \item otherwise, if no input character is available anymore, end the calculation.
\end{enumerate}

\item {\bf Transition:} \label{calc_I/O-A_transition} With $q^*$ as current state value and $i^*$ as current input character select a transition $t=(i^*, o^+, q^*, q^+)\in T^*$ and so determine $o^+$ and $q^+$. If there is no possible transition at this point, terminate the calculation with an error.

\item {\bf Repetition:} Set $q^* = q^+$ and $o^* = o^+$ and jump back to \ref{calc_I/O-A_loop}
\end{enumerate}
\end{definition}

As can be seen from the error condition in the execution rule, a protocol must fulfill certain consistency conditions to make sense.

\begin{definition}
A protocol is named \dots
\begin{enumerate}
\item \dots {\it ''closed''} if all output states are connected to some input state and no further input state is left.
\item \dots {\it ''well formed''} if each input character determined in step 2 can be processed in step 3.
\item \dots {\it ''interruptible''} if each interaction chain remains finite.
\item \dots {\it ''accepting''} if for each run the acceptance condition is fulfilled.
\end{enumerate}
A protocol that is closed, well formed, interruptible, and accepting is named {\it ''consistent''}.
\end{definition}

Keeping a protocol's consistency is also a condition that a renaming of I/O-characters in our engineering language has to fullfil. This fits nicely to the ''consistency management'' Johann van Bentham refers to in \cite{vBentham2008_Games}. 
\subsubsection{Example: The single-track railway bridge}
\begin{figure}[htbp]
\centering
\includegraphics[width=6cm]{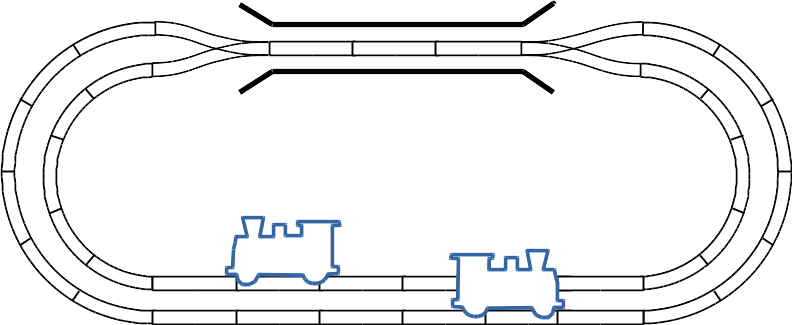} 
\caption{A single-track railway bridge crossed by two trains. To avoid a collision on the bridge, both trains interact with a central controller.}
\label{fig_trains}
\end{figure}

To illustrate the protocol concept I give the simple example of a single-track railway bridge drawn from \cite{Alur2015_Principles}. As is shown in Fig. \ref{fig_trains}, two trains, ${\mathcal Z}_1$ and ${\mathcal Z}_2$, must share the common resource of a single-track railway bridge. For this purpose, both trains interact with a common controller ${\cal C}$, which must ensure that there is no more than one train on the bridge at any one time.

\begin{figure}[h!]
\centering
\includegraphics[width=8cm]{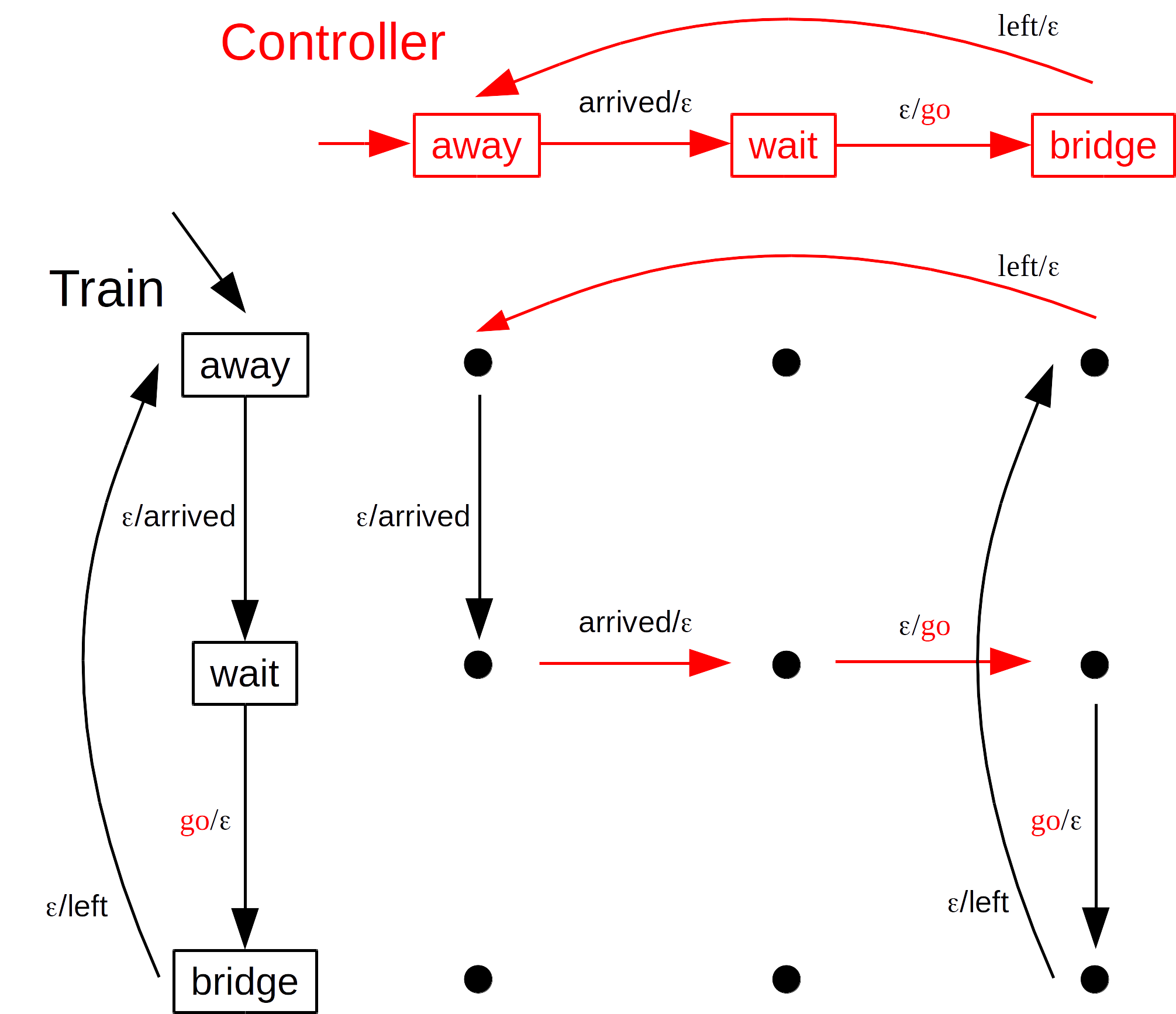}
\caption{In the protocol between train and controller for the problem of the single-track railway bridge, both controller and train are initially in the $away$ state. When a train arrives, it signals $arrived$ to the controller. This sign must now be processed by the controller, the controller in turn changes to its $wait$ state. The controller releases the track with $go$ and the train signals the controller with $away$ that it has left the bridge again.}
\label{fig_trains_protocol}
\end{figure}

The interaction between each train and the controller is modelled by a consistent protocol, illustrated in Fig. \ref{fig_trains_protocol}. For both the train and the controller we choose a model of 3 state values, which we call $Q_{Z_{1,2}/C}=\{away, wait, bridge\}$ for each train as well as for the controller. The input alphabet of the trains $I_{{\cal Z}_{1,2}} = \{go\}$ is the output alphabet of the controller $O_{\cal C}$ and the output alphabet of the trains $O_{{\cal Z}_{1,2}} = \{arrived, left\}$ is the input alphabet $I_{\cal C}$ of the controller. Fig. \ref{fig_trains_protocol} shows how the interaction via a Shannon state restricts the transition relation of the product automaton.

Intuitively, the train's state values represent the state of the train and the controller state values represent the train state as it is known to the controller. Please note that the correctness, we could also say the truth of the representation of the state of the train in the controller depends on the truthfullness of the train's signaling and the correctness of the protocol.

%
\section{Interaction semantics}\label{s_interaction_semantics}
%
Now we come to the semantics of an interaction. The idea is to use the mathematical concept of meaning with respect to formal languages also for interactions. As I outline in the appendix \ref{ss_concept_of_meaning_in_formal_languages}, the mathematical concept of attributing meaning to a term of a formal language consists of two steps: First an assignment of values to the variables of the term and then the application of an interpretation function to all of its elements. 
To apply this concept of meaning to the interaction between systems, we therefore have to identify the possible variables and the possible interpretation function of the interaction. 

Looking at our system + protocol concepts, we see that a state function can be viewed as assigning characters to a state variable at each point in time. With this understanding the representation of the system behavior by some I/O-TS of def. \ref{def_IO-transition_system_mit_epsilon} becomes the representation of the variable-value-assigment. 

And the interpretation function? In the world of systems, we can view the system function as ''interpreting'' the assignment by mapping one assignment onto the ''next'' one. Thus, the ''meaning'' of the assigned values are again assigned values. 
This is a major difference to the application of the meaning concept in formal languages. There, the interpretation function maps between different domains, namely terms and concepts. Here, the interpretaion function maps within a single domain, namely informatically denoted state values or 'characters'

However, in the world of interactions, we still have an issue: the transition relation of a protocol is usually nondeterministic and therefore does not define a function that maps assignments to assignments. There would be the possibility to look at sets of state values instead of single state values, as every nondeterministic transition relation does define a function, mapping the input character and internal state value onto a set of pairs of possible output characters and new state values. But this would not be in sync with the conception we have developed within the system notion.

\subsection{Decisions} \label{ss_entscheidungen}
The idea to stay within our interaction conception is to complete the nondeterministic transition relation of a protocol such that it becomes deterministic. I call the necessary additional input characters ''decisions''. In this sense we can say that decisions determine the behavior, that is, they determine those transitions which would otherwise be indeterminate \cite{Reich2023_VI_2}. 

Decisions in this sense complement the input alphabet $I$ of an I/O-TS ${\mathcal A}$ according to Def. \ref{def_IO-transition_system_mit_epsilon} to $I' = I\times D$ such that a complemented transition relation $\Delta'$ becomes deterministic.
In contrast to ordinary input characters, whose main characteristic is to appear in other output alphabets and that are allowed to appear in several transitions, we name all decisions of a corresponding transition system differently and different from all input and output characters and internal state values, so that we can be sure that they really do determine all transitions.	

\begin{definition} \label{def_decision} 
Be ${\mathcal A}$ an I/O-TS and $D$ an alphabet. The transition system ${\mathcal A}'$ is called a ''decision system'' or ''game in interactive form (GIF)''\footnote{Actually this ''game'' still lacks the utility function, a game in the traditional game theoretic sense has. But such a utility function is just one way to introduce a certain concept to determine decisions, which is, in this case, based on non-hierarchical, consistent preferences. There are many others.} to ${\mathcal A}$ and the elements of $D$ ''decisions'', if $\Delta' \subseteq (I_{\mathcal A}^\epsilon \times D) \times O_{\mathcal A}^\epsilon \times Q_{\mathcal A}\times Q_{\mathcal A}$  is the smallest possible set such that $I\cap D = \emptyset$, $O\cap D = \emptyset$, $Q\cap D = \emptyset$, and with $((i, d), o, p, q) \in \Delta'$ if $(i, o, p, q)\in \Delta$ and for $d$ applies:

\[
  d=\begin{cases}
    \epsilon, & \text{if there's no further transition $(\hat{i}, \hat{o}, \hat{p}, \hat{q}\in \Delta$}\\
              & \text{with $(i,p) = (\hat{i}, \hat{p}$.}\\
    \mbox{so selected} & \text{that $\Delta'$ is deterministic, i.e. $\Delta'$ determines the function}\\
              & \text{$f': I^\epsilon \times D^\epsilon \times Q \rightarrow O^\epsilon \times Q$ with $(o, q) = f'(i, d, p)$ such that }\\
              & \text{for two transitions $t'_1, t'_2 \in \Delta'$ holds $t'_1 \neq t'_2 \Rightarrow (d_1 \neq d_2$} \\
              & \text{or $d_1 = d_2 = \epsilon$).}\\
  \end{cases}
\]
\end{definition}

For a GIF we can modify the protocol execution rule \ref{calc_I/O-A_transition} {\it Transition} such that the selection choice becomes determined by some possible decision. 

Obviously, the set of decisions for an already deterministic I/O-A is empty. In Fig. \ref{fig_trains_decision_automaton} I illustrate the decision notion with the train-controller protocol. To determine the actions of train and controller three decisions are necessary. The train has to decide when to arrive and when to leave (''IArrive'' and ''ILeave'') and the controller has to decide when it let the train go (''ILetYouGo''). 

An execution fragment of a GIF is like the execution fragment of the protocol, but extended by the additional decisions.
\subsection{Meaning within an interaction}
The transition relation $\Delta_{\mathcal P}$ of a GIF ${\mathcal P}$ defines per construction a transition function $\delta:I_{\mathcal P}\times Q_{\mathcal P} \rightarrow O_{\mathcal P} \times Q_{\mathcal P}$. 

With this transition function of the GIF, we have an interpretation function we can use to define the meaning of an exchanged character $i$ with respect to some start value $p$ and possibly some decision $d$ --- which is the new state value $q$ with the possibly generated output character $o$:

\begin{equation}
(o,q) = \delta((i,d), p) =: interp_{d,p}(i) \label{eq_interpretation_interaction}
\end{equation}

A run $r$ is then the result of an interpretation of the initial assignment, that is the initial state $q_o$ and some input sequence of decisions $seq$: $r = \delta_{\mathcal A}^*(q_0, seq)$.

With this concept of interpretation we can define a fulfilment relation similar to that in appendix \ref{ss_concept_of_meaning_in_formal_languages} where we relate what we assign, some initial state $q_0$ together with a sequence of decisions $seq$, to the behaviour of the structure with the variables, the GIF. 

\begin{definition}
Let ${\mathcal P}$ be a GIF. We say that some initial state $q_0$ together with a sequence of decisions $seq$ fulfils ${\mathcal P}$, in symbolic notation $(q_0, seq) \models {\mathcal P}$, iff $interp_{q_0, seq}({\mathcal P})$ is an accepted run of the GIF.
\end{definition} 

\subsection{The meaning of a decision}
The interpretation function of a GIF defines the resulting state and output value as the meaning of a decision and an input character. However, if we are only interested in what happens by a decision until the next decision is taken, we can view all states that are reached by the ensuing interaction ping pong as being equivalent and lump them together by constructing equivalence classes. 

This leads us to the notion of $\epsilon$-closure in the decision space and to construct a somehow ''reduced'' decision automaton in a procedure similar to $\epsilon$-elimination for determining a deterministic from a nondeterministic finite automaton. 

I first define the $\epsilon$-decision closure of a state value $q$ as the set of all states that are accessible from $q$ without further decisions including $q$ itself:

\begin{definition}
Let ${\mathcal P}$ be a GIF. Then $h_\epsilon(q) = \{p\in Q\mid\, p = q \text{ or: if there is a } p'\in h_\epsilon(q) \, \text{ and there exists } i\in I \,\text{ and } o\in O^\epsilon \text{ s.t. } (i, \epsilon, o, p', p) \in \Delta_{{\mathcal P}}\}$ is the $\epsilon$-decision closure of the state value $q$.
\end{definition}

Please remember that if the decision is $\epsilon$, such an input character always exists. We can now attribute to every decision an $\epsilon$-decision closure as a ''target state'', which I name the {\it ''abstract meaning''} of the decision in contrast to its concrete meaning as it is directly provided by the GIF's transition function. With these sets of state values we can construct a quotient decision automaton as following: 

\begin{definition} \label{def_reduced_decision_automaton}
Let ${\mathcal P}$ be a GIF. Then I call the automaton ${\mathcal B}$, constructed by the following rules, the ''quotient decision automaton'' or ''game in its decision form (GDF)'' of ${\mathcal P}$:
\begin{itemize}
\item $Q_{\mathcal B}$ is the set of $\epsilon$-decision-closures of ${\mathcal P}$ that partitions $Q_{\mathcal P}$.
\item The input alphabet $I_{\mathcal B} = D_{\mathcal P}$ is the set of all decisions of ${\mathcal P}$.
\item The transition relation $\Delta_{\mathcal B} \subseteq I_{\mathcal B}\times Q_{\mathcal B}\times Q_{\mathcal B}$ is defined by: $(d, p, q) \in \Delta_{\mathcal B}$ if and only if there exists a reachable $p'\in p$ and some $i\in I_{\mathcal P}$, $o\in O_{\mathcal P}$, and $q'\in q$ such that $(i, d, o, p', q') \in \Delta_{\mathcal P}$.
\item The acceptance component $Acc_{\mathcal B}$ is defined by: $p\in Q_{\mathcal B}$ is an element of $Acc_{\mathcal B}$ if and only if at least one $p' \in p$ is an element of $Acc_{\mathcal P}$.
\item The initial state ${q_0}_{\mathcal B}$ is defined as ${q_0}_{\mathcal B} = h_\epsilon({q_0}_{\mathcal P})$. 
\end{itemize}  
\end{definition}

\begin{figure}[h!]
\centering
\includegraphics[width=8cm]{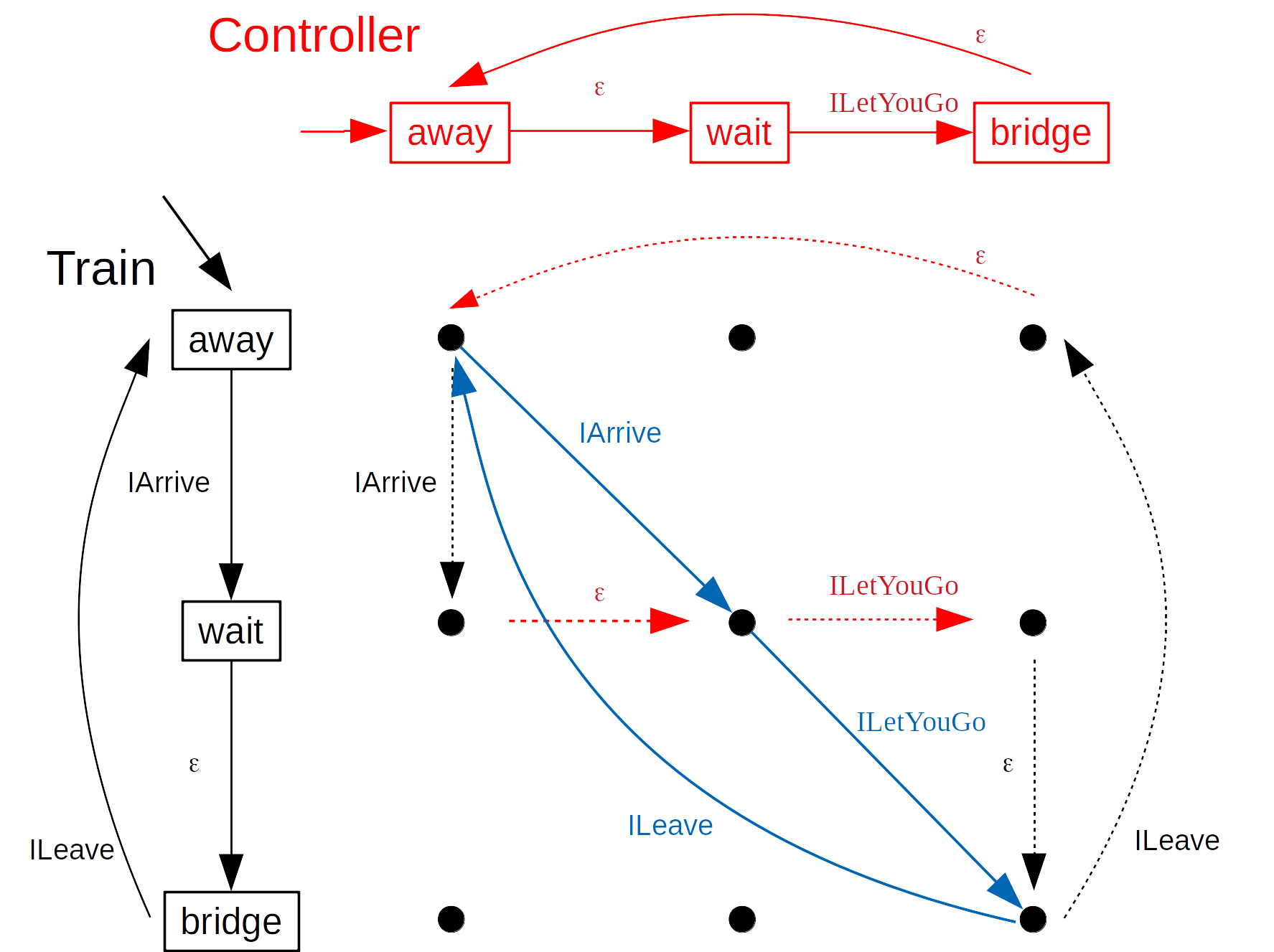}
\caption{The transitions of the GDF of the train-controller GIF are shown in blue. The train decides that it arrives and leaves (''IArrive'' and ''ILeave'') and the controller decides when it let the train go (''ILetYouGo'')}
\label{fig_trains_decision_automaton}
\end{figure}

The resulting reduced decision automaton or GDF abtracts from interaction as the input- and output characters do no longer appear. It is deterministic with $D$ as its input alphabet. Please note, that it relates to the product state space of the GIF (or protocol). I illustrate the transformation of a GIF to a GDF in Fig. \ref{fig_trains_decision_automaton} for the train-controller GIF. 

As the states of a GDF are the abstract meanings of the decisions of a GIF and a GDF is more or less a game in the sense of traditional game theory, I think that it is justified to say that, in our sense, game theory is the theory of the meaning of decisions.

We can go a step further and ask, which decisions have the same meaning? Which we translate in: which decisions make a GDF to transit into one concrete  state value? The resulting equivalence classes of equivalent decisions correspond, in my opinion, to our intuitive understanding of decisions and their meaning. 

Intuitively, we judge decisions by their consequences rather independently of the initial situation. For example, whether a child hears a violin and decides ''to play the violin'', or whether it sees a picture and decides ''to play the violin'': in both cases we would say that the child takes the same decision. 

Now, we can formulate this more precisely that both decisions are different but have the same meaning, as both result in the same state of the child to play the violin and therefore are, in this sense, equivalent.

\subsection{The meaning of an exchanged character} \label{s_abstract_meaning_of_character}
Now, we tackle our initial problem: what is the meaning of an incoming character? Looking at our GIF, the significance of the incoming characters is their role as a coupling mechanism which enables each participant to take the possible decisions according to the rules of the interaction-game. 

Because any exchanged character may occur in several transitions of an interaction (in contrast to decisions), we cannot define their meaning without referring to the set concept. Also, we have to take into account the starting state of their transitions.
In other words, the feature of semantic ambiguity is very deeply rooted in our model of interaction semantics.

So, while the concrete meaning of an incoming character is given be the interpretation function of a GIF according to equation \ref{eq_interpretation_interaction}, to abstract we can look for the decisions an incoming character enables. These are its complementary selection decisions, or -- in case there was none (it was only $\epsilon$) -- any directly ensuing spontaneous decision. I therefore define

\begin{definition} \label{def_abstract_meaning_of_character}
Let ${\mathcal P}$ be a GIF and let $i\in I$ occur in all transitions $(i,d_j,*,p,q_j) \in \Delta$ for a given $p$ with $d_j$ for $j=1,\dots, n$. If $n > 1$ then these $n$ selection decisions $d_j$ are the elements of the select. If $n = 1$ then all directly ensuing spontaneous decisions are the elements of the select. I call the set of the abstract meanings of the associated select-decisions the potential meanings of $i$ with respect to $p$.
\end{definition}   

It appears to me that by defining the abstract meaning of a character we somehow aribitrarily have to restrict ourselves to some order of reachability, where definition \ref{def_abstract_meaning_of_character} grasps the significance of a character only to ''first order''.

Based on this definition, it is clear that two characters $i_1$ and $i_2$ have the same potential meanings with respect to a state value $p$, if the application of the associated select decisions finally result in the same target state.

\subsection{Composition of meaning in the I-language}
Is the meaning of characters and decisions as we have defined it in the previous sections compositional? This means, that the semantics of a sequence of characters or decisions can be deduced solely from the semantics of the characters or decisions themselves. Mathematically, this translates into the following definition: 

\begin{definition} The meaning of two characters or decisions $c_1$, $c_2$ is compositional if an operator $op$ and a concatenation operation $.$ exists such that for the interpretation function holds $interp(c_1 . c_2) = op(interp(c_1), interp(c_2))$.
\end{definition}

From this definition we see that the compositionality of meaning is not given per se, but depends on the existence of a clear concatenation conception and on the existence of the function $op$.  

The first thing we have to clarify is what we mean be ''concatenation''. From the perspective of a single system, two incoming characters occur consecutively if one succeeds the other as input of the system. However, from an interaction perspective, two characters occur consecutively, if they are consecutively exchanged between two systems.  

The next thing we have to clarify is what our operator $op$ is supposed ''to know''. It knows actually only $ (o_1, q_1) = interp_{p_1}(c_1)$ and $(o_2, q_2) = interp_{p_2}(c_2)$. If both interpretations represent consecutive transitions, that is $p_2 = q_1$, then its clear that the output of our operator is just $(o_2, q_2)$. But if they do not represent consecutive transitions, then our operator cannot calculate the composed meaning.

Additinally, it is clear that the compositionality of the meaning of characters and decisions depends on the well-definedness of the transition relation. In settings where the transition relation, i.e. the interaction context itself, becomes the object of consideration, compositionality of meaning is also  lost. 

\subsection{Natural language semantics as a solution to the simultaneous interaction and understanding (SIAU)-problem}

The second key idea of this contribution is to realize how the approach of interaction semantics fits with the property of natural language semantics to separate the relevant from the irrelevant. 

Applying our model of semantics to the question how it comes that speaker and listener actually do understand each other, we found ourselves trapped in a chicken-and-egg problem: on the one hand, purposeful interaction with other speakers requires mutual understanding and, on the other hand, establishing mutual understanding requires purposeful interaction. 

For problems of this chicken-and-egg class, computer science offers other, by now well understood examples, such as the ''simultaneous localisation and mapping (SLAM)'' problem (e.g. \cite{ThrunBurgardFox2005}): On the one hand, to determine one's position in a terrain, I need a map. On the other hand, to determine the map, I need to know my own position. 

An obvious solution for such chicken-and-egg problems is iterative: an internal model is increasingly improved through constant contact with the environment. Following the acronym SLAM, I propose to speak of the ''simultaneous interaction and understanding (SIAU)'' problem. 

There exists extensive research where the iterative character of acquiring knowledge about interaction semantics is already investigated. This could be on an evolutionary time scale (e.g. \cite{BenzEbertJaegervanRooij2011} for a brief overview) or on an online-timescale. An example for the latter is Sida I. Wang, Percy Liang and Christopher D. Manning \cite{WangLiangManning2016} who explore the idea of language games in a learning setting, which they call interactive learning through language game (ILLG). The objective is to transform a start state into a goal state,  but the only action the human can take is entering an utterance.

It's the requirement of internal modelling that makes it understandable why semantics in natural language, at its core, distinguishes the relevant from the irrelevant. The internal model cannot be created ''photographically'', with all possible recordable details, but has to be limited to what is recognised as ''essential'' for reasons of efficiency. This internal modelling thus requires distinguishing the relevant from the irrelevant -- and can therefore be, accordingly, both appropriate or inappropriate.  

\begin{figure}
\centering
\includegraphics[width=3.5in]{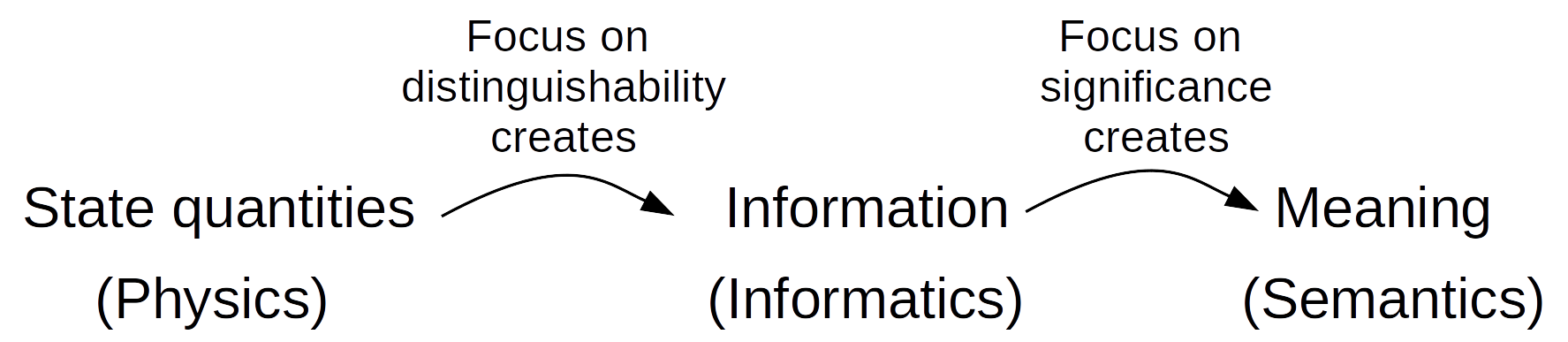}
\caption{The relation of the concepts of physical state quantities, information and meaning.}
\label{fig_physics_informatics_semantics}
\end{figure}

In summary, we can relate physics, informatics and semantics in the sense illustrated in Fig. \ref{fig_physics_informatics_semantics}: We introduce informatics by focussing on the distinguishable and we introduce natural language semantics by focussing on the relevant. 

%
\section{Similar other work} \label{s_other_work}      
%
As I have already indicated in the beginning, this article touches scientific, engineering as well as philosophical aspects. Here I only present other approaches that are somewhat similar to mine.

Carolyn Talcott \cite{Talcott1997} uses the term ''interaction semantic'' of a component to denote the set of sequences of interactions in the sense of input or output messages or silent steps in which it might participate. She  composes her components of multisets of so called actors with unique addresses where the actor semantics could be either internal transitions as a combination of execution and message delivery steps or interaction steps with an exchange of messages. 
In her formalism she takes into account that the interaction semantics must be invariant against renaming of addresses, state and message values but she neither addresses any semantic fulfillment relation nor the concept of the ''meaning'' of a single exchanged character, not to mention decisions. 
In summary, her approach is very similar to the $\pi-calculus$ \cite{Milner1992} but her addresses refer to actors with state and not to stateless channels and the interactions are asynchronous.

Tizian Schröder and Christian Diedrich \cite{SchroederDiedrich2020_Formal} also view the semantics of the exchanged characters within an interaction as being provided by their processing and use the same model of a discrete system with its system function $f$ as processing model. 
They also define decisions to determine a unique result for the system function in otherwise nondeterministic interactions. But according to my understanding, their model rests on the assumption of the system function being a bijection. 

\subsection{Dialogical logic} 
Mathematicians have also addressed the relation between meaning, knowledge, and logic in the context of interactions in the sense of games or dialogues under the notions of ''dialogical logic'' \cite{LorenzenLorenz1978} or ''game-theoretical semantics'' \cite{HintikkaSandu1997}. The former focuses on real human discourse while the latter focuses on model-oriented analysis of the logical meaning of linguistic sentences and its relation to certain rule-governed human activities. The basic idea of Hintikka's evaluation game is that as a proof, a Verifier tries to find a winning strategy in a two person game against a Falsifier such that a given first order formula $\phi$ is true in a given Model ${\mathcal M}$ under some assignment of the variables. Negation, conjunction and disjunction are translated into role switches and choice attributions. 
 
\subsection{Computational semantics} 
Also related is the field of computational semantics as it is concerned with computing approximations of the meanings of linguistic objects such as sentences, text fragments, and dialogue contributions (e.g. \cite{BuntMuskens1999,Boleda2020_DistributionalSemantics}).

%
\section{Discussion} \label{s_discussion}     
%
The aim of this article was to delineate a model of interaction semantics and en passant provide a concrete understanding of the meaning of characters within an interaction. 

Its key idea was to apply the mathematical concept of meaning, as it was developed to provide meaning to formal expressions, to interactions. I therefore argue that with the same intuition as we talk about the semantics of a formal language we can talk about the meaning of an interaction. 

I showed that the proposed concept of semantics does indeed convey the essential aspects of what we colloquially denote as semantics, as my model 
\begin{enumerate}
\item is based on the model of information transport and processing;
\item clarifes the relationship to models of formal semantics;
\item accounts for the fact that meaning is invariant against appropriate renaming;
\item locates the concept of meaning in our usual technical description of interactions;
\item defines when two different characters have the same meaning;
\item defines what is an ''interpretation'' and what is an ''interpretation context''; and
\item explains under which conditions meaning is compositional, and when it is not.
\end{enumerate}

With my model, we do get a better understanding of a couple of important issues:
First, the tight relation between the informatical approach to model interactions and game theory. 
The traditional view on game theory is expressed by Roger B. Myerson \cite{Myerson1991_Game_Theory} who defines game theory as {\it ''the study of mathematical models of conflict and cooperation between intelligent rational decision-makers''}. 

Using game theory to analyze the conventional character of language and determine any conception of meaning already requires a very elaborate conceptual framework. We have to know what a game is, what decisions are, how they relate to our physical world, whether a signal is an action or something else, etc. And, even more importantly, traditional games, neither in their strategic nor in their extensive form account for any information exchange, that is interaction. 

David Lewis tried to fill this gap with signaling games, but this remains a rather ad hoc extension. With our approach we now can say that game theory abstracts from interaction and is indeed the theory of the meaning of decisions. 

Second, the ’simultaneous interaction and understanding (SIAU)’ problem. The proposed model provides a conceptual framework, where we better comprehend that understanding and interacting are both creative acts, that therefore are amenable to an iterative, interrelated process designed to continually expand the interpretive subject's scope of action.  

Perhaps, the real power of the meaning concept in informatics will unfold its full potential only when we start to tackle the SIAU problem with iterative algorithms. In analogy to the solution of the SLAM problem, this probably will requires us to explore how to represent easy context identification and switching capabilities and language-expressible vague knowledge properly structured to improve it iteratively, based on the speaker's experience. And not in the sense of a ''speech-collage'' were a system learns how to formulate sentences in a way that it becomes difficult to distinguish them from those generated by a knowledgeable system.

I think, that the proposed semantic model also argues against the traditional distinction between semantics and pragmatics \cite{Morris1938_Theory_of_Signs} as it puts the change of state in the center and identifies the meaning of our exchanged signs with their use. 

Third, why ontologies are less ’semantic’ then their proponents suggest. Ontologies were introduced in the 1980s as a modeling tool and as a component of knowledge-based systems. According to Thomas A. Gruber \cite{Gruber1995}, an ontology is an ''{\it explicit specification of ... the objects, concepts, and other entities that are assumed to exist in some area of interest and the relationships that hold among them}'', that is, an explicit specification of a conceptualization in the sense of \cite{GeneserethNilsson1987}. C. Feilmayr and W. Wöß \cite{FeilmayrWoess2016} claim along this tradition, that an ontology-based application design allows ''{\it a clear and machine-interpretable basis for meaning}''.

But in fact, ontologies rest on a mapping model of semantics between different domains. And therefore their semantics is that of a formal language, based on semantic atoms (its basic concepts) together with composition rules. And very often, these basic concepts are not clearly defined  \cite{FeilmayrWoess2016}. Actually, there is not even a sufficient agreement on the concept of ontology itself, for instance in distinction to taxonomy \cite{FeilmayrWoess2016}. Martin Hepp formulates \cite{Hepp2008_Ontologies} ''{\it In fact, it is a kind of paradox that the seed term of a novel field of research, which aims at reducing ambiguity about the intended meaning of symbols, is understood and used so inconsistently}''.

And forth, we better understand how ’semantic’ interoperability is to be achieved. Actually, from an interaction theoretic perspective, ''semantic interoperability'' is a pleonasm, comparable to a ''free gift''. Interoperability by definition refers to the mutually coordinated processing of information from different systems, which -- then again by definition -- constitutes precisely the semantic aspect of interactions.

In this sense, as already stated in section  \ref{ss_informatics_and_semantics}, everything that relates to information processing is somehow ''semantic''. With this understanding, object orientation \cite{Jacobson1992} is nothing else than fully specifying the semantics of some information. In contrast, data typing can be understood as specifying the processing of information to a lesser degree, namely that we know only in general how to process the respective information in the sense that we know some atomic operations together with the rules of computability for their composition -- but not how they will be processed concretely \cite{Reich2018_Data,Bitkom2020_Klassifikation_Interaktionen}. 

Also, the well known layered architecture models of software engineering capture semantic aspects, as long as they are based on consistent structures of information processing \cite{Reich2021_Komposition_Interoperabilitaet}

For the standardization of interoperability, this understanding of semantics has far-reaching consequences. It puts the 'semantic'\footnote{in contrast to 'transport' protocols.} protocols in the center. With their closedness, semantic protocols like SEPA or SS7 set the frame what terms need to be understood in what level of detail by the participants. Without such a frame, any standardization effort is severely hampered by the undefined semantic exploration space. 

I would like to conclude this article with some speculative thoughts. From a philosophical point of view, the model of interaction semantics implies that without interpretation, the world is meaningless. Concretely it says (or means) that meaning is attributed by interpretation and without such a mapping which we declare as interpretation, there is no meaning. More abstractly it says that the notion of meaning depends necessarily on the notion of interpretation performed by some interpreter and there is no meaning in any absolute sense. 

Based on the presented concept of meaning, one could speculate that the ''flow of thought'' we introspectively experience when we consciously think is based on ''anticipated interactions'' which would bind our capability to think abstractly reciprocally and thereby tightly to our ability to express ourselves language-wise, just as our ability to imagine playing an instrument like a violin depends on the extend of practice on this instrument.

And finally, one could further speculate that sense and sensibility are inseparable if we understand our sensibility as a mode of understanding. If we are calm or angry, if we hate or love, we essentially interpret our world differently. Our emotions modulate our intuition about what is relevant or not (see e.g. \cite{ShackmanLapate2018} for an overview, how emotion and cognition interact).

{\bf Acknowledgments:} I specially thank Tizian Schröder and Christian Diedrich with whom I had many fruitful discussions on this topic some time ago.


\bibliographystyle{vancouver}
\bibliography{informatics,soziologie,philosophy}

%
\section{Appendix A: The concept of meaning in formal languages} \label{ss_concept_of_meaning_in_formal_languages}
%
To create the concept of interaction semantics, the central idea of this work is to apply the procedure model of formal semantics to a suitable interaction model. I therefore now present this model of semantics of formal languages in more detail.

Based on Alfred Tarski \cite{Tarski1935}, formal languages are structured according to a certain scheme. First, the syntax is defined, consisting of a set of allowed characters together with a set of rules describing which expressions can be formed. Then, in a second step, the semantics is defined by an interpretation function determining the meaning of the allowed expressions by mapping them to entities that are legitimately assumed to exist and about we can talk in our normal language. 

I illustrate this briefly with the example of the propositional calculus. For this we assume that we already know what a proposition is, that a proposition can be either true or false and that we can state at least some elementary propositions. Then, the calculus describes how one can obtain further statements from elementary statements by $and$, $or$ and $negation$ operations. In order to ensure the distinction between the expressions attributed to the calculus and those attributed to our normal language, I put all calculus expressions in quotation marks.

The colloquial expressions that we use to formulate the rules of syntax and semantics deserve special attention. To formulate propositional logic, we have to use so called ''propositional forms''. Syntactically, propositional forms correspond to calculus expressions, but they belong to our engineering language as we use them to define calculs expressins with special variables as placeholders for real calculus expressions. I write down propositional forms like calculus expressions in quotation marks, but symbolize the special variables with a prefixed \$ sign, in order to be able to distinguish them reliably from the variables which are part of the calculus. 
 
The allowed characters of the propositional calculus are determined by the alphabet $\{''\!w'', ''\!f''\}$, the set of operator characters $\{''\!\vee'', ''\!\wedge'', ''\!\neg''\}$, as well as the set of characters representing variables for propositions $V = \{''\!p'', ''\!q'', etc.\}$.

The syntax rules for building propositions are: 
\begin{enumerate}
  \item $''\!w''$ and $''\!f''$ are propositions;
  \item Each variable is a proposition;
  \item Are $''\!\$a''$ and $''\!\$b''$ propositions, then $''\!\neg \$a''$, $''\!\$a \vee \$b''$ and $''\!\$a \wedge \$b''$ are also propositions.
\end{enumerate}

The interpretation of a proposition $''\!\$a''$, ${\cal I}_{b}(''\!\$a'')$, provides its meaning and consists of 
\begin{enumerate}
  \item an assignment of truth values to all variables: $b:V\rightarrow \{true, false\}$, where $true$ and $false$ are expressions of our colloquial engineering language we hopefully fully comprehend. 
  \item a recursive rule that determines the meaning of the proposition:
  \begin{enumerate}
    \item ${\cal I}_b(''\!w'') = true$; ${\cal I}_b(''\!f'') = false$;
    \item ${\cal I}_b(''\!p'') = b(''\!p'')$;
    \item ${\cal I}_b(''\!\neg \$a'') = true[false]$ if ${\cal I}_b(\$a) = false[true]$;
    \item ${\cal I}_b(''\!\$a \vee \$b'') = true$, if ${\cal I}_b(\$a) = true$ $or$ ${\cal I}_b(\$b) = true$;
    \item ${\cal I}_b(''\!\$a \wedge \$b'') = true$ if ${\cal I}_b(\$a) = true$ $and$ ${\cal I}_b(\$b) = true$.
  \end{enumerate}
\end{enumerate}

Because the interpretation maps each syntactically correct formula to truth values, we can also define a ''fulfillment''-relation $\models$ where an assignment $b$ fulfills a formula $''\!\$a''$, or, symbolically, $(b, ''\!\$a'') \in \models$, or, in the usual infix notation, $b \models ''\!\$a''$ iff ${\cal I}_b(''\!\$a'') = true$.  

One interesting aspect of the semantics of a formal language is its homogenously compositional character, that is that the meaning of composed terms results exclusively from the meaning of its parts.

\end{document}